\documentclass[letterpaper, 10 pt, conference]{ieeeconf}  

\IEEEoverridecommandlockouts                              

\usepackage{cite}
\usepackage{amsmath}
\usepackage[caption=false,font=footnotesize]{subfig}
\usepackage{amssymb}
\usepackage{amsfonts}
\usepackage{tabularx}
\usepackage{booktabs}
\usepackage{mathtools}
\usepackage{graphicx}
\usepackage{placeins}
\usepackage{siunitx}
\usepackage{float}
\DeclareMathOperator*{\argmin}{arg\,min}
\makeatletter
\@ifundefined{NAT@parse}{}{\let\NAT@parse\undefined}
\makeatother
\usepackage[hidelinks]{hyperref}
\hypersetup{
  draft=false,        
  bookmarks=false,    
  hypertexnames=false,
  colorlinks=true,    
  linkcolor=magenta,
  citecolor=magenta,
  urlcolor=magenta
}

\title{\LARGE \bf
U-LAG: Uncertainty-Aware, Lag-Adaptive Goal Retargeting for Robotic Manipulation
}

\author{Anamika J H\textsuperscript{*}, Anujith Muraleedharan\textsuperscript{*}%
\thanks{\textsuperscript{*}Equal contribution.}%
\thanks{This work was not supported by any organization.}%
}

\begin{document}

\maketitle
\thispagestyle{empty}
\pagestyle{empty}

\begin{abstract}

Robots manipulating in changing environments must act on percepts that are late, noisy, or stale. We present U-LAG, a mid-execution goal-retargeting layer that leaves the low-level controller unchanged while re-aiming task goals (pre-contact, contact, post) as new observations arrive. Unlike motion retargeting or generic visual servoing, U-LAG treats in-flight goal re-aiming as a first-class, pluggable module between perception and control. Our main technical contribution is UAR–PF, an uncertainty-aware retargeter that maintains a distribution over object pose under sensing lag and selects goals that maximize expected progress. We instantiate a reproducible Shift×Lag stress test in PyBullet/PandaGym for pick, push, stacking, and peg insertion, where the object undergoes abrupt in-plane shifts while synthetic perception lag is injected during approach. Across 0–10 cm shifts and 0–400 ms lags, UAR–PF and ICP degrade gracefully relative to a no-retarget baseline, achieving higher success with modest end-effector travel and fewer aborts; simple operational safeguards further improve stability. Contributions: (1) UAR–PF for lag-adaptive, uncertainty-aware goal retargeting; (2) a pluggable retargeting interface; and (3) a reproducible Shift×Lag benchmark with evaluation on pick, push, stacking, and peg insertion.
\end{abstract}

\section{INTRODUCTION}
Manipulation pipelines increasingly run with asynchronous perception, so observations can arrive late or intermittently and invalidate precomputed contact goals mid-trajectory. Setpoints that were safe at plan time can be stale when the end effector reaches them, causing missed grasps, glancing pushes, and wasted motion. Classical visual servoing closes the loop directly on image/feature error and is highly effective at high control rates, but it degrades under sensing delay or occlusion \cite{c1,c2}. At the other end of the spectrum, task-and-motion planning (TAMP) commits to waypoints and repairs or replans when the world changes \cite{c3,c4,c5}; and receding-horizon MPC reacts more quickly, but still recomputes trajectories that can already be out-of-date when perception lags \cite{c6,c7}. Learning-based visuomotor policies \cite{c8,c9} and grasp-detection modules integrated with servo control \cite{c10,c11} absorb small perturbations, and recent trends in large-scale data and diffusion-based policies push reactivity further \cite{c12,c13}, yet most systems still assume timely observations or fold all adaptation into the low-level controller.
\begin{figure}[!t]
  \centering
  \includegraphics[
    width=\columnwidth,
    height=0.40\textheight,
    keepaspectratio
  ]{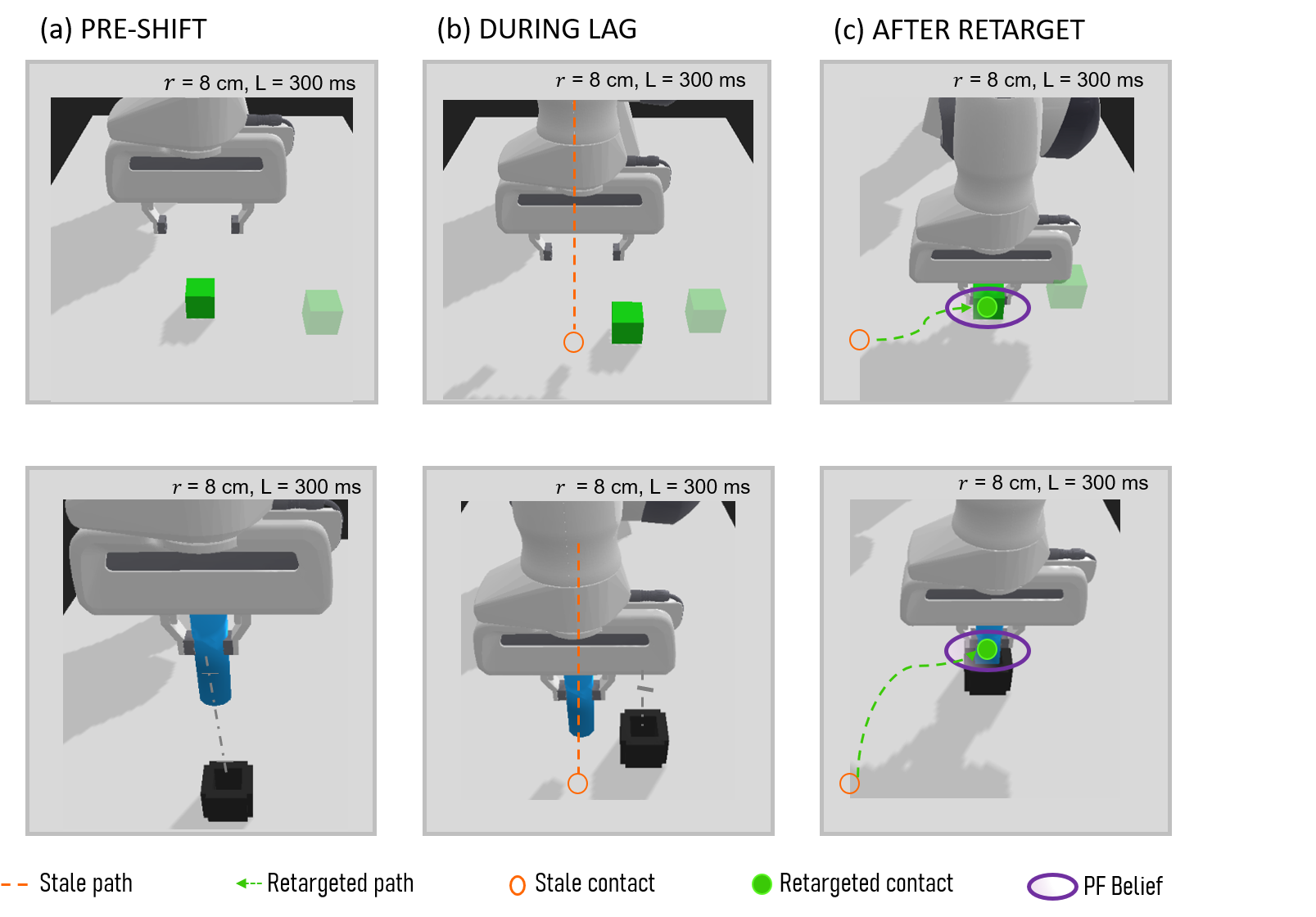}
  \vspace{-0.5em}
  \caption{\textbf{U-LAG in action under Shift$\times$Lag.}
  Snapshots for pick, push, stacking, and peg insertion at
  $(r{=}8\,\mathrm{cm},\,L{=}300\,\mathrm{ms})$.
  Dashed orange: stale target; green: updated goal; purple ellipse: contact region.}
  \label{fig:teaser}
  \vspace{-0.9em}
\end{figure}
We advocate a separation of concerns: treat mid-execution goal retargeting as its own module between perception and control. Rather than redesigning controllers or replanning from scratch, a retargeting layer updates pre-contact, contact, and post-contact goals whenever new observations arrive, while an unchanged Cartesian servo executes those goals. This viewpoint is complementary to visual servoing and planning—it revises targets instead of commanding velocities from pixels, and it drops into existing stacks with minimal integration effort.

\begin{figure*}[t]
  \centering
  \includegraphics[height=0.31\textheight,keepaspectratio]{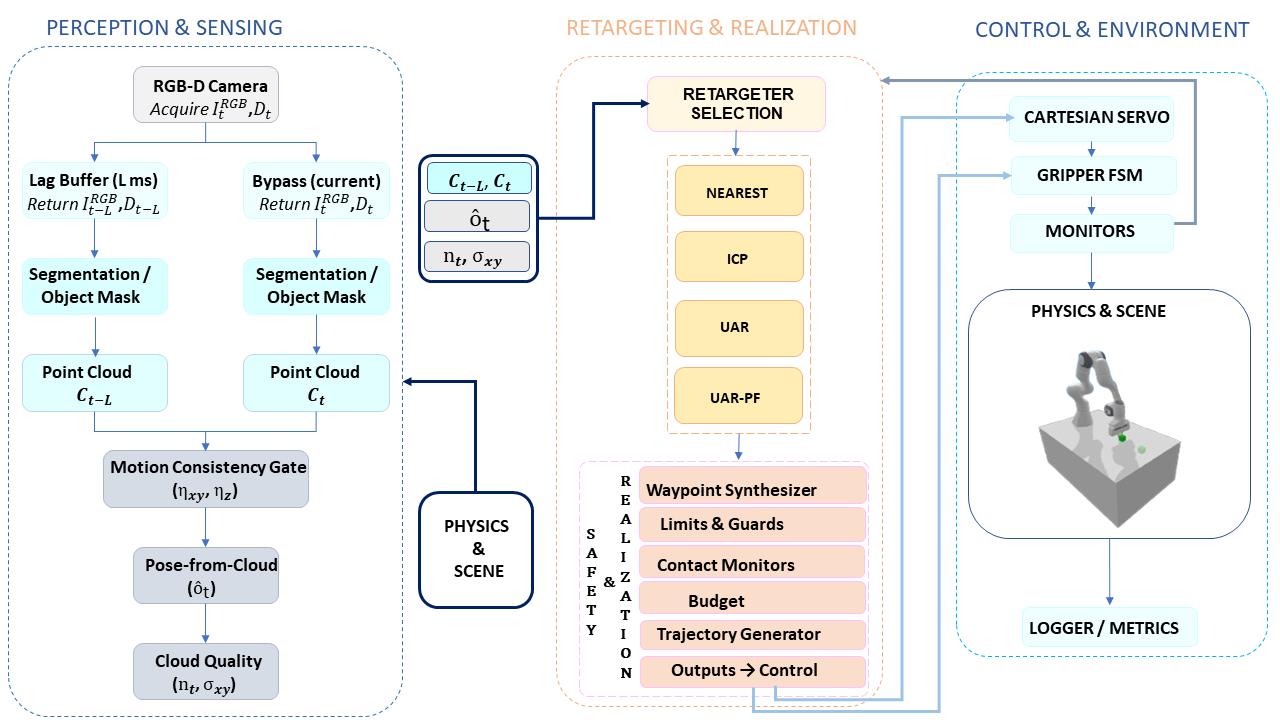}
  \vspace{-0.6em}
  \caption{\textbf{U-LAG system overview.} Left: perception \& sensing. Center: retargeters (Nearest/ICP/UAR/UAR--PF), shared guards and a task-specific waypoint synthesizer. Right: unchanged Cartesian servo \& gripper FSM executing waypoints.}
  \label{fig:ulag-arch}
  \vspace{-0.8em}
\end{figure*}

We instantiate this idea as U-LAG (Uncertainty-Aware, Lag-Adaptive Goal Retargeting): a lightweight layer that consumes RGB--D observations and outputs refreshed task goals during execution. Within a common interface, we realize multiple retargeters, including deterministic nearest-geometry updates, geometric registration to reconcile earlier and current observations, and a particle-filtered variant that maintains a belief over object pose under delay \cite{c14}. An overview is shown in Fig.~\ref{fig:ulag-arch}. Lightweight reliability checks draw on uncertainty cues from perception to guard against outliers while preserving the pluggable abstraction\cite{c15}.

To probe delay and disturbance systematically, we introduce a Shift×Lag protocol that injects controlled object shifts at approach time and holds actions against lagged goals for prescribed durations before the retargeter updates the setpoints. We evaluate across pick-and-place, pushing, stacking, and peg insertion, spanning non-prehensile and contact-rich regimes represented in prior  \cite{c8, c16, c17, c18}. Across shifts (0–10 cm) and lags (0–400 ms), U-LAG variants degrade gracefully relative to a no-retarget baseline; registration- and particle-filter–based realizations remain robust in the harder regimes.

\section{Related Work}
\subsection{Visual Servoing and Reactive Policies}
Visual servoing directly couples image measurements with robot motion and has been studied extensively\cite{c1,c2}. Both image-based and position-based formulations have been applied to grasping and assembly tasks\cite{c19,c20}. While effective at high control rates, classical servoing is sensitive to sensing delays and occlusions. Recent extensions replace hand-crafted features with deep descriptors or learned keypoints\cite{c21,c22}, and predictive/preview variants forecast image features to mitigate delay\cite{c23,c24}. Learning-based visuomotor policies trained via reinforcement learning\cite{c8,c9}or grasp-detection networks integrated with servoing\cite{c10,c11} improve robustness to perturbations. More recent methods use large-scale data or diffusion models for reactive manipulation and policy learning\cite{c12,c13}. Despite these advances, most systems assume timely perception updates; performance degrades when state estimates lag behind execution.

\subsection{Planning and Mid-Execution Replanning}
Task-and-motion planning (TAMP) combines symbolic reasoning with geometric constraints to produce feasible manipulation sequences \cite{c3,c4}. Feedback TAMP updates goals during execution by repairing waypoints as the world state changes\cite{c5}, while push–grasping policies adapt local goals online in cluttered scenes\cite{c16}. For motion generation, sampling-based planners compute collision-free trajectories in high-dimensional spaces\cite{c25}, and contact-implicit methods optimize trajectories that reason over intermittent contact \cite{c26}. To react under uncertainty, Model Predictive Control (MPC) repeatedly optimizes over short horizons\cite{c6,c7}; stochastic/MPPI variants and belief-space planners incorporate uncertainty explicitly in the optimization\cite{c27,c28}. Recent approaches also integrate learned dynamics and differentiable optimization to accelerate replanning\cite{c29,c30}. Although powerful for global robustness, these methods often recompute entire trajectories when perception lags, leaving the currently executing setpoints stale; motivating a lightweight goal-retargeting layer like U-LAG.
\subsection{Uncertainty in Perception and Manipulation}
Uncertainty modeling has been explored in depth estimation \cite{c15}, optical flow \cite{c31}, and VO/SLAM \cite{c32}. Learned confidence scores improve robustness in challenging conditions \cite{c33}. In manipulation, probabilistic filters track object pose under noise \cite{c14}, while Bayesian grasp detection accounts for sensor uncertainty \cite{c10}. Recent work investigates uncertainty-aware keypoint selection \cite{c34}and contact-rich manipulation with stochastic models \cite{c35}. In robotic vision, large-scale models have introduced uncertainty estimates for depth and segmentation \cite{c36}. Most work targets state estimation rather than goal retargeting.
\subsection{Task-Specific Adaptation: Pushing, Stacking, and Peg Insertion}
For pushing, push–grasp synergies and self-supervised interaction policies achieve robust recovery in clutter \cite{c16}, while classical push–grasp formulations handle contact planning and non-prehensile moves \cite{c37}, and stochastic/learned contact models enable precise control under contact uncertainty \cite{c35}. For stacking, large-scale visuomotor RL and grasp-then-place pipelines are widely used robustness testbeds \cite{c8}. For peg insertion, classical compliance-based assembly established precision under uncertainty \cite{c17}, followed by learning-based insertion with deep RL that adapts to pose and friction errors \cite{c18}. Across these domains, prior systems typically specialize control or invoke replanning; in contrast, U-LAG provides a task-agnostic, mid-execution goal-retargeting layer applicable across pick-and-place, pushing, stacking, and insertion under perception delay and scene change.

\section{METHODOLOGY}

\subsection{Problem Formulation}
\label{subsec:problem}

We consider a fixed-base manipulator interacting with a rigid object whose pose at time $t$ is
$\mathbf{o}_t = [x_t,\,y_t,\,z_t]^\top \in \mathbb{R}^3$ in a world frame $\mathcal{W}$.
A task specifies a goal setpoint $\mathbf{g}$---for \emph{pick}, a grasp pose; for \emph{push}, a target object position in the table plane. Perception provides a depth image and a segmented point cloud
$C_t=\{\,\mathbf{p}_k \in \mathbb{R}^3\,\}$,
but with latency $L$\,ms: at control time $t$ the most recent available observation is $C_{t-L}$.
During the approach phase, the object may undergo an abrupt in-plane perturbation
$\mathbf{s}\in\mathbb{R}^2$ (e.g., external contact), making the precomputed targets stale.

\paragraph*{Retargeting layer}
We insert a pluggable \emph{goal retargeting} module
\begin{equation}
\label{eq:retarget-map}
\mathcal{R}:\; \big(C_{t-L},\,C_t,\,\text{state}\big)\;\mapsto\;
\big(\mathbf{g}^{\mathrm{pre}},\,\mathbf{g}^{\mathrm{contact}},\,\mathbf{g}^{\mathrm{post}}\big),
\end{equation}
which updates the \emph{pre-contact}, \emph{contact}, and \emph{post} waypoints consumed by the (unchanged) low-level controller.
We realize $\mathcal{R}$ with four alternatives: ICP alignment, Nearest-geometry recomputation, Uncertainty-Aware Retargeting (UAR), and UAR with particle filtering (UAR--PF), each wrapped by reliability mechanisms (Sec.~\ref{subsec:reliability}).

\paragraph*{Assumptions}
(i) quasi-static interaction over a control tick; (ii) single rigid grasp/push contact;
(iii) calibrated camera with a known extrinsic to $\mathcal{W}$; and
(iv) a reliable object segment in simulation (any detector/segmenter can provide it in practice).
Orientation is not critical for pushing, while for picking we use a standard parallel-jaw grasp pose; both are handled within the waypoints produced by Eq.~\eqref{eq:retarget-map}.

\subsection{Sensing \& State Estimation}
\label{subsec:sensing}
A calibrated depth camera provides a segmented point cloud
$C_t=\{\mathbf{p}_k\in\mathbb{R}^3\}$ at control time~$t$, with perception latency~$L$ (so the freshest usable frame may be $C_{t-L}$). We use $C_t$ when available; otherwise we fall back to the most recent cloud.

\subsubsection{Point-Cloud Pose Proxy}
\label{subsubsec:poseproxy}
We estimate the object center $\hat{\mathbf{o}}_t=[\hat{x}_t,\hat{y}_t,\hat{z}_t]^\top$ from cloud geometry rather than a full 6D pose. We trim depth outliers using percentile clipping and form a robust inlier set
\[
\tilde{C}_t=\{\mathbf{p}\in C_t \mid \mathrm{perc}_1(p^z)\le p^z \le \mathrm{perc}_{99}(p^z)\}.
\]
Let $\boldsymbol{\ell}_t=\min_{\mathbf{p}\in\tilde{C}_t}\mathbf{p}$ and
$\mathbf{h}_t=\max_{\mathbf{p}\in\tilde{C}_t}\mathbf{p}$; the AABB center is
$\hat{\mathbf{c}}_t=\tfrac{1}{2}(\boldsymbol{\ell}_t+\mathbf{h}_t)$.
With object size $a$ (cube edge), we set $\hat{z}_t=h_t^z-\tfrac{a}{2}$ and
$\hat{\mathbf{o}}_t=[\hat{c}_t^x,\hat{c}_t^y,\hat{z}_t]^\top$.
For pushing we fix a top-down orientation; for picking we use a standard parallel-jaw grasp pose.
If $|\tilde{C}_t|$ is too small, the proxy is marked invalid and we guard/skip updates
(Sec.~\ref{subsec:reliability}).

\subsubsection{Cloud Quality Metrics}
\label{subsubsec:cloudquality}
We summarize cloud reliability with two scalars:
(i) inlier count $n_t=|\tilde{C}_t|$ and
(ii) planar dispersion $\sigma_{xy}=\sqrt{\lambda_{\max}(\mathrm{cov}\{[p^x,p^y]^\top:\mathbf{p}\in\tilde{C}_t\})}$.
Optionally we record depth spread $\sigma_z=\operatorname{std}\{p^z\}$.
These terms (a) scale UAR/UAR--PF margins and (b) feed a \emph{vision guard}:
if $\|\hat{\mathbf{o}}_{t,xy}-\hat{\mathbf{o}}_{t-L,xy}\|>\tau_{xy}$ or
$|\hat{z}_t-\hat{z}_{t-L}|>\tau_z$, we ignore the fresh cloud and fall back to a trusted estimate
(Sec.~\ref{subsec:reliability}). We also perform an early re-sense and a brief peek-lift before recapture to reduce occlusions.

\subsubsection{Peg insertion sensing}
We ray-cast to collect a \emph{rim} point cloud (top-of-wall hits) around the socket. The socket center is the AABB midpoint of rim hits; if none are observed we fall back to the median of interior table returns. The rim height $\hat{z}_{\mathrm{rim}}$ is the median rim $z$ (or a nominal value if unavailable). Each trial uses exactly two scans (pre-shift and post-lag).
\subsection{Goal Geometry}
\label{subsec:goal-geometry}

\noindent\textit{Pick.}
Given the sensed proxy (Sec.~\ref{subsec:sensing})
$\hat{\mathbf{o}}_t=[\hat{x}_t,\hat{y}_t,\hat{z}_t]^\top$, the retargeting layer emits a pre-grasp hover, a grasp pose, and a post-lift. The grasp height is set using the top-of-object plane plus a small clearance, and orientation follows a standard parallel-jaw grasp for the object.

\medskip
\noindent\textit{Push.}
Let $\mathbf{c}_{xy}=[\hat{x}_t,\hat{y}_t]^\top$ and $c_z=\hat{z}_t$ from $\hat{\mathbf{o}}_t$.
For a desired table-plane goal $\mathbf{g}_{\mathrm{obj}}=[x^\star,y^\star]^\top$, define
\begin{equation}
\hat{\mathbf{d}} \;=\;
\frac{\mathbf{g}_{\mathrm{obj}} - \mathbf{c}_{xy}}
     {\left\| \mathbf{g}_{\mathrm{obj}} - \mathbf{c}_{xy} \right\| + \varepsilon},
\qquad \varepsilon>0.
\end{equation}
Let $a$ denote the cube edge, and let $(\delta_{\mathrm{pre}},\delta_{\mathrm{contact}},
\delta_{\mathrm{over}},\delta_z)$ be the pre-contact standoff, contact margin, post overshoot, and vertical offset. The XY waypoints are
\begin{align}
\mathbf{g}^{\mathrm{pre}}_{xy}
&= \mathbf{c}_{xy} - \hat{\mathbf{d}}\!\left(\tfrac{a}{2} + \delta_{\mathrm{pre}}\right), \\
\mathbf{g}^{\mathrm{contact}}_{xy}
&= \mathbf{c}_{xy} - \hat{\mathbf{d}}\!\left(\tfrac{a}{2} + \delta_{\mathrm{contact}}\right), \\
\mathbf{g}^{\mathrm{post}}_{xy}
&= \mathbf{g}^{\mathrm{contact}}_{xy} +
     \hat{\mathbf{d}}\!\left(\left\| \mathbf{g}_{\mathrm{obj}} - \mathbf{c}_{xy} \right\|
     + \delta_{\mathrm{over}}\right),
\end{align}
with push height $z_{\mathrm{push}} = c_z + \delta_{z}$ and 3D targets
$\mathbf{g}^{\mathrm{pre}}=[\mathbf{g}^{\mathrm{pre}}_{xy}{}^\top,\, z_{\mathrm{push}}]^\top$,
$\mathbf{g}^{\mathrm{contact}}=[\mathbf{g}^{\mathrm{contact}}_{xy}{}^\top,\, z_{\mathrm{push}}]^\top$,
$\mathbf{g}^{\mathrm{post}}=[\mathbf{g}^{\mathrm{post}}_{xy}{}^\top,\, z_{\mathrm{push}}]^\top$.

\medskip
\noindent\textit{Stacking (place target).}
Let the base cube’s top AABB have center $(x_g,y_g)$ and top height $z_{\text{top}}$.
With cube edge $a$, the place target for the held cube is
\[
\mathbf{g}_{\text{stack}} \;=\; [\,x_g,\; y_g,\; z_{\text{top}} + a/2\,]^\top,
\]
and success/tuning use $(\mathrm{tol}_{xy},\,\mathrm{tol}_z)$.

\medskip
\noindent\textit{Peg insertion (insert target).}
Let $\hat{z}_{\mathrm{rim}}$ be the sensed rim height and $z_{\mathrm{table}}$ the table top.
We move to a pre-insert hover
\[
z_{\mathrm{pre}} \;=\; \hat{z}_{\mathrm{rim}} + h_{\mathrm{over}},
\]
then descend while clamping $(x,y)$ to the socket center. The final insert depth target is
\[
z_{\mathrm{center}} \;=\; z_{\mathrm{table}} + \tfrac{1}{2}h_{\mathrm{peg}} + 4~\mathrm{mm},
\]
executed by a simple Z-axis feedback insert.
\subsection{Retargeters \texorpdfstring{$\mathcal{R}$}{R}}
\label{subsec:retargeters}

Given delayed/fresh clouds $(C_{t-L},C_t)$ and the controller state,
the retargeting layer $\mathcal{R}$ outputs updated waypoints
$\big(\mathbf{g}^{\mathrm{pre}},\mathbf{g}^{\mathrm{contact}},\mathbf{g}^{\mathrm{post}}\big)$,
to be consumed by the fixed low-level Cartesian servo. We instantiate four variants.

\subsubsection{Nearest}
\label{subsubsec:nearest}
Recompute $\big(\mathbf{g}^{\mathrm{pre}},\mathbf{g}^{\mathrm{contact}},\mathbf{g}^{\mathrm{post}}\big)$
directly from the latest sensed proxy $\hat{\mathbf{o}}_t$ using the closed-form
construction in Sec.~\ref{subsec:goal-geometry}. This is deterministic and fast,
and performs strongly when clouds are clean and lags are modest.

\subsubsection{ICP Retargeting (ICP)}
\label{subsubsec:icp}
Align the stale and fresh clouds with point-to-point ICP and transport only the stale
pre-contact target through the estimated rigid transform. Let
$\{ \mathbf{p}_i\}\!\subset\!C_{t-L}$ and $\{ \mathbf{q}_{\pi(i)}\}\!\subset\!C_{t}$ be paired correspondences; we solve
\begin{equation}
\label{eq:icp}
\mathbf{R}^\star,\mathbf{t}^\star \;=\;
\argmin_{\mathbf{R}\in\mathrm{SO}(3),\,\mathbf{t}\in\mathbb{R}^3}
\sum_{i} \big\| \mathbf{R}\,\mathbf{p}_i + \mathbf{t} - \mathbf{q}_{\pi(i)} \big\|^2,
\end{equation}
with standard outlier rejection and early stopping.
We then set
\begin{equation}
\mathbf{g}^{\mathrm{pre}\,\prime} \;=\; \mathbf{R}^\star \, \mathbf{g}^{\mathrm{pre}} + \mathbf{t}^\star,
\qquad
z'_{\mathrm{push}} \;=\; \hat{z}_t + \delta_z,
\end{equation}
and recompute $\big(\mathbf{g}^{\mathrm{contact}},\mathbf{g}^{\mathrm{post}}\big)$
from the latest proxy $\hat{\mathbf{o}}_t$ (Sec.~\ref{subsec:goal-geometry}).
ICP excels when the overlap between $C_{t-L}$ and $C_t$ is high.

\subsubsection{Uncertainty-Aware Retargeting (UAR)}
\label{subsubsec:uar}
Inflate geometric margins according to cloud dispersion to bias against glancing contact
and under-reach. With XY dispersion $\sigma_{xy}$ (Sec.~\ref{subsubsec:cloudquality}),
define an inflation term $\Delta(\sigma_{xy}) = \lambda\,\sigma_{xy}$ and use
\begin{align}
\delta_{\mathrm{pre}}'     &= \mathrm{clip}\!\left(\delta_{\mathrm{pre}} + \Delta,\; \underline{\delta}_{\mathrm{pre}},\, \overline{\delta}_{\mathrm{pre}}\right), \\
\delta_{\mathrm{contact}}' &= \mathrm{clip}\!\left(\delta_{\mathrm{contact}} + \Delta,\; \underline{\delta}_{\mathrm{contact}},\, \overline{\delta}_{\mathrm{contact}}\right), \\
\delta_{\mathrm{over}}'    &= \mathrm{clip}\!\left(\delta_{\mathrm{over}} + \tfrac{1}{2}\Delta,\; \underline{\delta}_{\mathrm{over}},\, \overline{\delta}_{\mathrm{over}}\right),
\end{align}
while $z$ uses $c_z+\delta_z$. Here $\Delta(\sigma_{xy}) \equiv \lambda\,\sigma_{xy}$ inflates the nominal offsets; we then clip to task-specific bounds before substituting into Sec.~\ref{subsec:goal-geometry}. We treat the clamps $\underline{\delta}_{(\cdot)}$ and $\overline{\delta}_{(\cdot)}$ as fixed per task, where $(\cdot)\in\{\mathrm{pre},\mathrm{contact},\mathrm{over}\}$; the numeric bounds are listed in Table~\ref{tab:params}. The updated offsets are substituted into Sec.~\ref{subsec:goal-geometry} to produce waypoints.

\subsubsection{Uncertainty-Aware Retargeting with Particle Filtering (UAR--PF)}
\label{subsubsec:uar--pf}
We maintain a particle filter \cite{c38} with $N$ particles over the planar state and heading
$\mathbf{s}=[x,\,y,\,\theta,\,v_x,\,v_y,\,\omega]^\top$ with weights $\{w_i\}$. At the first valid proxy, we sample each particle independently from the same Gaussian $\mathcal{N}([\hat{x}_0,\hat{y}_0,\hat{\theta}_0]^{\!\top},\,\operatorname{diag}(\sigma_{0,xy}^2,\sigma_{0,xy}^2,\sigma_{0,\theta}^2))$ and set uniform weights.
\paragraph*{Prediction (CV with Gaussian accel noise).}
Let $\mathbf{p}=\begin{bmatrix}x\\y\\\theta\end{bmatrix}$,
$\mathbf{v}=\begin{bmatrix}v_x\\v_y\\\omega\end{bmatrix}$, and
$\mathbf{a}=\begin{bmatrix}a_x\\a_y\\\alpha\end{bmatrix}$. With controller period $\Delta t$,
\begin{align}
\mathbf{v} &\leftarrow \mathbf{v} + \mathbf{a}\,\Delta t, \\
\mathbf{p} &\leftarrow \mathbf{p} + \mathbf{v}\,\Delta t .
\end{align}
Acceleration noise is
\begin{equation}
a_x,a_y \sim \mathcal{N}(0,\sigma_v^2), \qquad
\alpha \sim \mathcal{N}(0,\sigma_\omega^2).
\end{equation}

\paragraph*{Measurement update (planar center).}
Given proxy $\mathbf{z}=\hat{\mathbf{c}}_{xy}$ and dispersion $\sigma_{xy}$,
\begin{equation}
w_i \propto \exp\!\Big(-\tfrac{1}{2}\,\big\lVert \mathbf{c}_i-\mathbf{z}\big\rVert_{\Sigma^{-1}}^{2}\Big),
\label{eq:pf-like}
\end{equation}
where $\mathbf{c}_i=[x_i,\,y_i]^\top$ and $\Sigma=\sigma_{xy}^2 I_2$.
If the cloud is empty or guarded at this tick, we skip the update (prior only).

\paragraph*{Resampling.}
Systematic resampling when
\begin{equation}
\mathrm{ESS}=\frac{1}{\sum_i w_i^2} \;<\; \frac{N}{2},
\end{equation}
after which $w_i\!\leftarrow\!1/N$.

\paragraph*{Waypoint generation with UAR inflation.}
The mean $\bar{\mathbf{c}}_{xy}=\sum_i w_i [x_i,y_i]^\top$ seeds the geometry in
Sec.~\ref{subsec:goal-geometry}; the push height uses the current vertical proxy $c_z$
(the PF tracks planar pose only). We apply the same inflation $\Delta=\lambda\sigma_{xy}$
(Sec.~\ref{subsubsec:uar}) to standoff, contact, and overshoot clamps before forming
$(\mathbf{g}^{\mathrm{pre}},\mathbf{g}^{\mathrm{contact}},\mathbf{g}^{\mathrm{post}})$.

\subsection{Reliability mechanisms (shared)}
\label{subsec:reliability}

We include practical wrappers that do not change the low-level controller or the retargeters, but improve robustness and reproducibility. They are \emph{not} claimed as algorithmic contributions.

\paragraph{Vision guard.}
Given a fresh proxy $\hat{\mathbf{o}}_t$ and the previous (or delayed) proxy $\hat{\mathbf{o}}_{t-L}$, if
$\lVert \hat{\mathbf{o}}_{t,xy}-\hat{\mathbf{o}}_{t-L,xy}\rVert > \tau_{xy}$ or
$|\hat{z}_t-\hat{z}_{t-L}| > \tau_z$,
we ignore the fresh cloud and fall back to a trusted pose (PF prior on hardware; ground truth only in simulation).
Unless stated otherwise, we set $\tau_{xy},\tau_z$ large enough that the guard was effectively disabled unless the cloud was clearly corrupted.

\paragraph{Two-stage waypointing (UP$\rightarrow$XY$\rightarrow$DOWN).}
Any retargeted pose $\mathbf{g}_{\text{new}}=[g_x,g_y,z_{\text{push/grasp}}]^\top$ is realized as three waypoints:
$[x_t,y_t,z_{\text{safe}}] \rightarrow [g_x,g_y,z_{\text{safe}}] \rightarrow [g_x,g_y,z_{\text{push/grasp}}]$,
with $z_{\text{safe}}=\max(z_t+\Delta_z,\ \hat{z}_t+a/2+\text{clear})$.
Arrival tolerances are $(\epsilon_{xy},\epsilon_z) \approx (15\,\text{mm},\,10\,\text{mm})$.
This prevents lateral sweeps near the surface and reduces accidental contact.

\paragraph{Slip monitor.}
During push, we query contact proximity between the end-effector and object. If contact is lost for $>N_{\text{lost}}$ consecutive control ticks before reaching the goal, we trigger a single re-sense/replan when the replan budget permits (no replans for the no-retarget baseline).

\paragraph{Stall monitor.}
If both EE motion and goal progress remain below small thresholds over a time window while contact is maintained (classic jam), we trigger the same re-sense/replan; if the replan budget is exhausted, we abort the trial. This avoids unbounded thrashing near the goal.

\paragraph{Settings.}
Unless stated, we use $(\Delta_z,\ \text{clear})=(2\,\text{cm},\,1\,\text{cm})$,
$(\epsilon_{xy},\epsilon_z)=(15\,\text{mm},10\,\text{mm})$,
$N_{\text{lost}}=8$ ticks, stall window $0.4$\,s with velocity and progress thresholds $<1$\,mm/tick.

\subsection{Control Interface}
\label{subsec:control}

The low-level controller is a first-order Cartesian position servo that accepts a per-tick delta command
$\Delta\mathbf{x}_t\!\in\!\mathbb{R}^3$ with a hard cap:
\begin{equation}
  \|\Delta\mathbf{x}_t\|_2 \le \Delta_{\max} \quad \text{(default } \Delta_{\max}\!=\!2.5\text{--}3~\mathrm{mm}\text{)}.
\end{equation}
At each control tick $t$, given the current target $\mathbf{g}_t$ (either the next queued waypoint or the final goal), we send
\begin{equation}
  \Delta\mathbf{x}_t \;=\; \operatorname{clip}_{\Delta_{\max}}\!\big(\mathbf{g}_t - \mathbf{x}_t\big),
\end{equation}
where $\mathbf{x}_t$ is the measured end-effector (EE) pose and $\operatorname{clip}_{\Delta_{\max}}(\cdot)$ scales the vector to satisfy the norm bound if necessary.
The controller integrates these capped delta commands toward the target, and we log cumulative end-effector travel for analysis as
$S \leftarrow S + \|\Delta\mathbf{x}_t\|_2$ at each control tick.

\paragraph*{Gripper.}
For push, the gripper remains open. For pick, the sequence is open $\rightarrow$ close $\rightarrow$ open at the pre-grasp, grasp, and post-lift subgoals, respectively.
\vspace{0.25em}

\begin{table}[t]
  \caption{Parameters and defaults used in all experiments.}
  \label{tab:params}
  \centering
  \footnotesize
  \setlength{\tabcolsep}{3pt} 
  \renewcommand{\arraystretch}{1.05}
  \begin{tabularx}{\columnwidth}{@{}p{0.36\columnwidth} X@{}}
    \toprule
    \textbf{Category} & \textbf{Default / Description} \\
    \midrule
    Object \& clearances
      & Cube edge $a=0.04$\,m; vertical clearance $1$\,cm. \\
    Waypoint offsets (push)
      & $\delta_{\text{pre}}=20$\,mm,\; $\delta_{\text{contact}}=4$\,mm,\;
        $\delta_{\text{over}}=10$\,mm,\; $\delta_z=0$\,mm. \\
    Two-stage tolerances
      & $\epsilon_{xy}=15$\,mm,\; $\epsilon_z=10$\,mm. \\
    Slip thresholds
      & $N_{\text{lost}}=8$ ticks; replan budget $=1$
        (\emph{no-retarget}: $0$). \\
    Stall detector
      & Window $0.4$–$0.6$\,s; EE/goal progress thresholds $\approx 1$–$2$\,mm. \\
    UAR
      & Margin slope $\lambda$ (tuned per task); uncertainty samples $M$. \\
    UAR–PF
      & Particles $N\!\in\![64,128]$; low-variance resample if $\mathrm{ESS}<N/2$. \\
    \bottomrule
  \end{tabularx}
\end{table}

\subsection{Discussion}
\label{subsec:discussion}
U\textendash LAG isolates \emph{goal retargeting} as a modular layer between perception and control. Nearest and ICP reflect two classical extremes (closed-form geometry versus alignment), UAR injects uncertainty awareness at essentially zero cost, and UAR--PF adds belief smoothing that is especially valuable under large lags and noisy clouds. Shared reliability mechanisms (guarding, two-stage waypointing, slip/stall) are the pragmatic engineering that turns retargeting into a robust behavior under contact and sensing imperfections; they complement but do not replace the core contribution---uncertainty-aware, lag-adaptive retargeting.

\section{Experimental Results}
\subsection{Setup and Protocol}

\paragraph{Environment}
We use PyBullet \cite{c39} with PandaGym (Franka Panda). A single fixed RGB--D pinhole camera (known intrinsics/extrinsics, $60^\circ$ FOV) provides depth and point clouds. Control and servo details follow Sec.~\ref{subsec:control}.

\paragraph{Tasks}
We evaluate four tasks—\emph{pick}, \emph{push}, \emph{stacking}, and \emph{peg insertion}. All use the same Shift$\times$Lag stressor (teleport + synthetic delay) with minor task-specific triggers noted where relevant. Quantitative results are reported per task.

\paragraph{Shift$\times$Lag stressor}
When the end-effector (EE) reaches the approach trigger, we (i) teleport the \emph{target object} by a random in-plane vector of fixed magnitude $r\in\{0,2,\dots,10\}$\,cm, then (ii) hold actions against the stale target for a synthetic perception delay $L\in\{0,100,200,300,400\}$\,ms. After the lag elapses, the camera re-captures and the retargeter updates the goals. All modes (except \emph{no-retarget}) perform exactly one post-lag retarget at the trigger; any later re-sense/replan events can only come from the shared slip/stall budget ($\le 1$). Retarget latency is measured from lag onset to the moment a new goal is selected (before motion).

\paragraph{Baselines / Retargeters}
We compare \emph{no-retarget}, \emph{Nearest}, \emph{ICP}, \emph{UAR}, and \emph{UAR--PF} as defined in Sec.~\ref{subsec:retargeters}.

\paragraph{Reliability wrappers (shared)}
We use the shared wrappers of Sec.~\ref{subsec:reliability}.

\paragraph{Grid / Trials \& Metrics}
We sweep $6$ shifts $\times$ $5$ lags with $50$ random seeds per cell. Each run logs success, abort, replans, EE travel, retarget latency, and final/min goal distance. We report success rate, abort rate, mean replans, mean EE travel, and mean retarget latency.

\paragraph{Runtime}
All retargeters execute in a few milliseconds at our cloud sizes: Nearest/UAR $\mathcal{O}(1)$; ICP $\mathcal{O}(I\,m)$ with $I\!\le\!30$; UAR--PF $\mathcal{O}(N)$. Overhead is negligible relative to injected $L$ and the 100\,Hz control loop.
\begin{figure*}[!t]
  \centering
  \setlength{\tabcolsep}{1.5pt}
  \renewcommand{\arraystretch}{0}
  \begin{tabular}{@{}ccccc@{}}
    \includegraphics[width=0.195\linewidth]{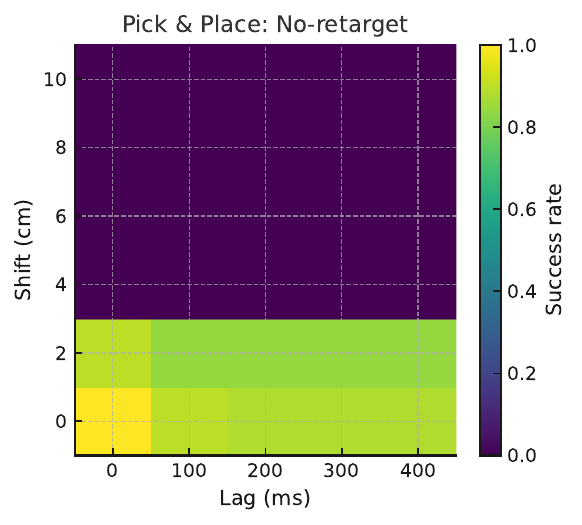} &
    \includegraphics[width=0.195\linewidth]{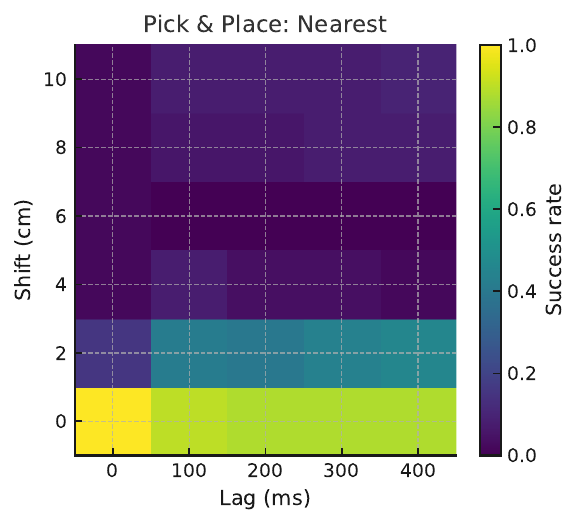} &
    \includegraphics[width=0.195\linewidth]{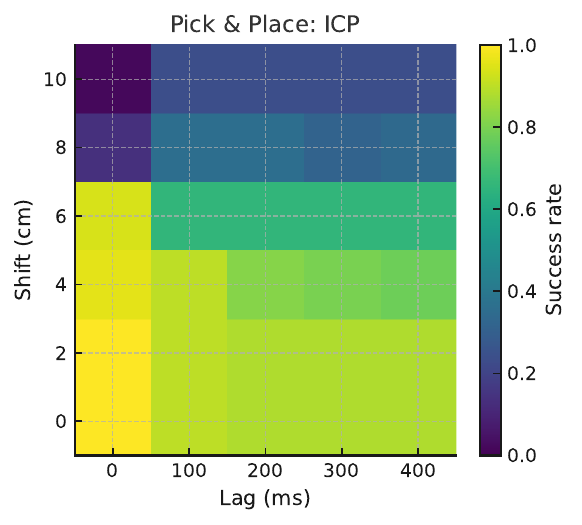} &
    \includegraphics[width=0.195\linewidth]{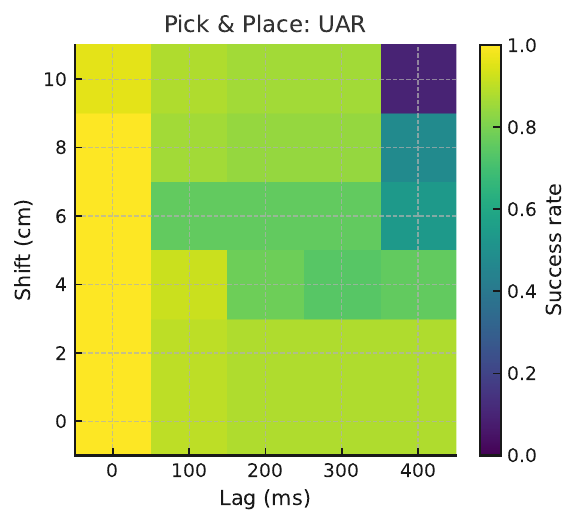} &
    \includegraphics[width=0.195\linewidth]{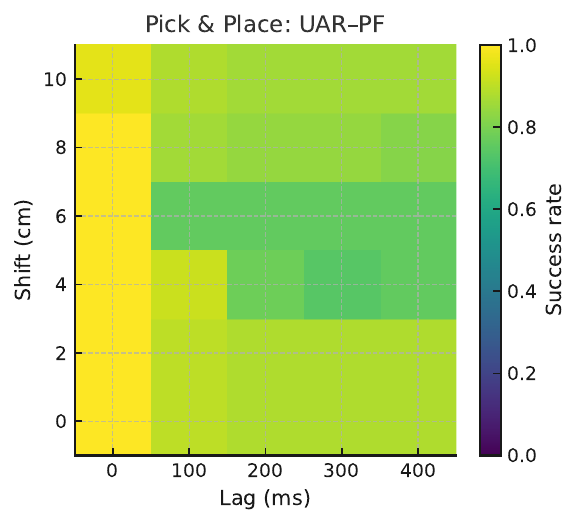} \\
    \includegraphics[width=0.195\linewidth]{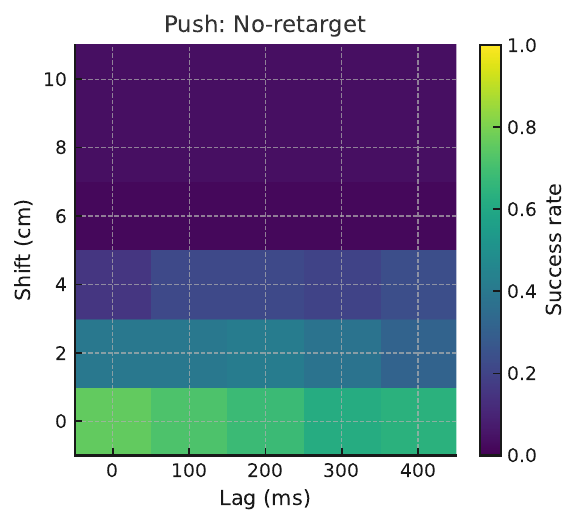} &
    \includegraphics[width=0.195\linewidth]{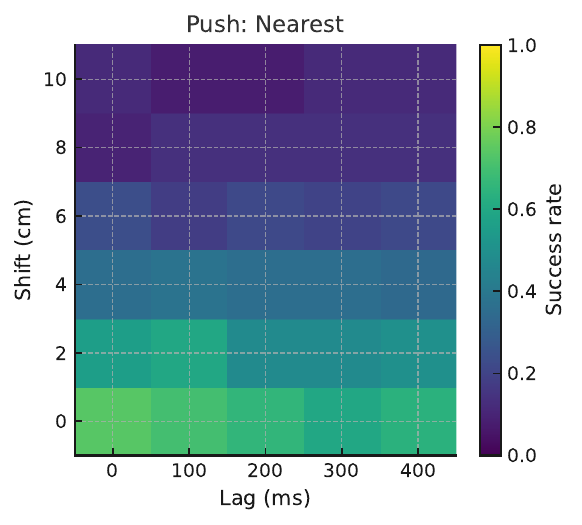} &
    \includegraphics[width=0.195\linewidth]{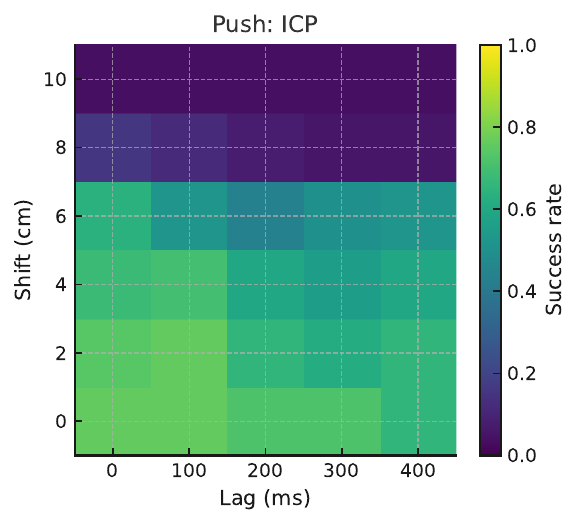} &
    \includegraphics[width=0.195\linewidth]{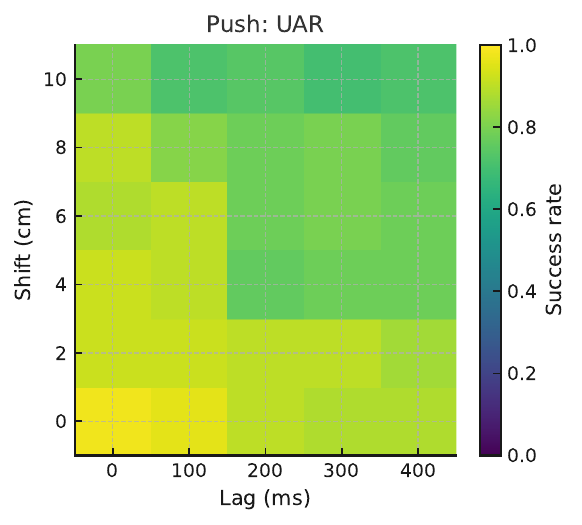} &
    \includegraphics[width=0.195\linewidth]{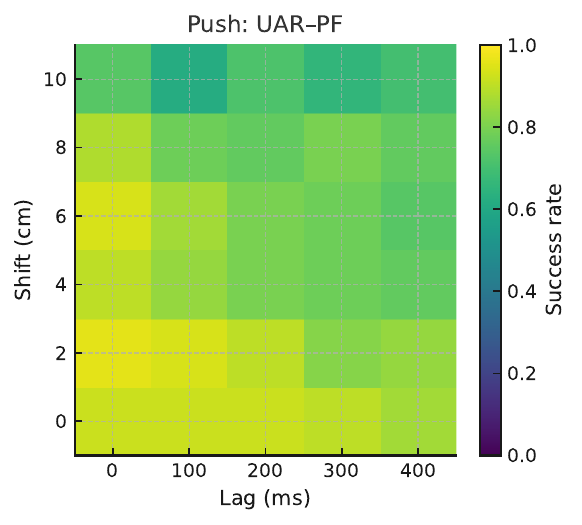} \\
    \includegraphics[width=0.195\linewidth]{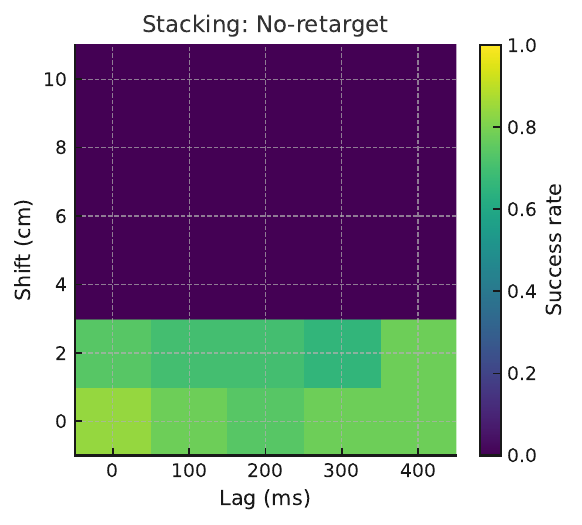} &
    \includegraphics[width=0.195\linewidth]{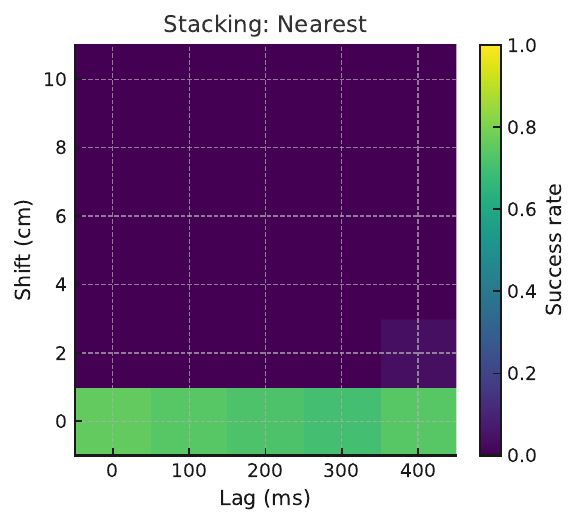} &
    \includegraphics[width=0.195\linewidth]{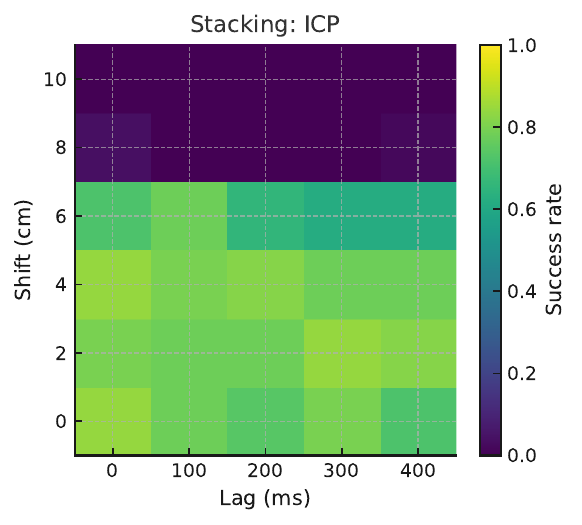} &
    \includegraphics[width=0.195\linewidth]{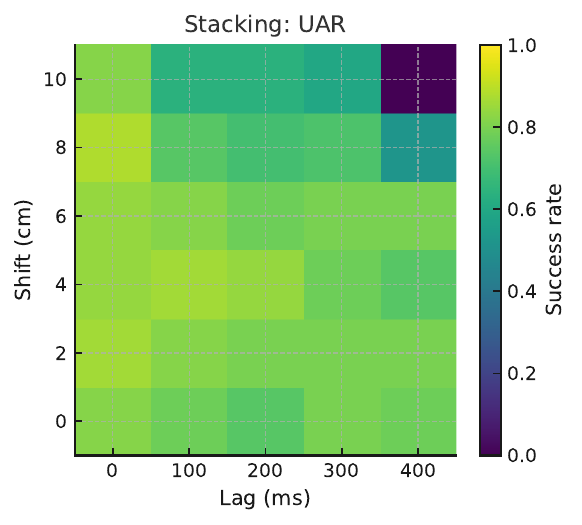} &
    \includegraphics[width=0.195\linewidth]{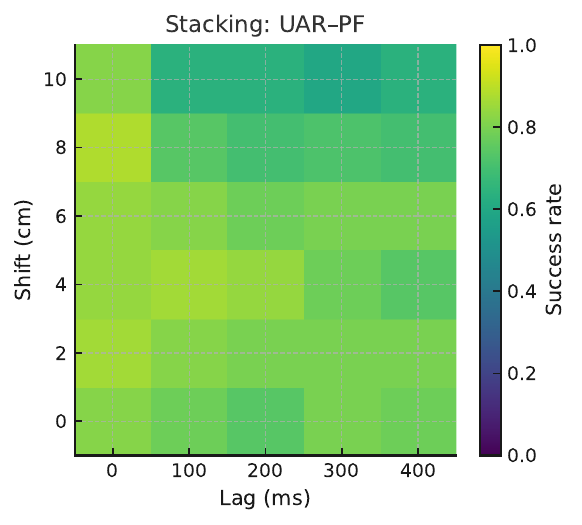} \\
    \includegraphics[width=0.195\linewidth]{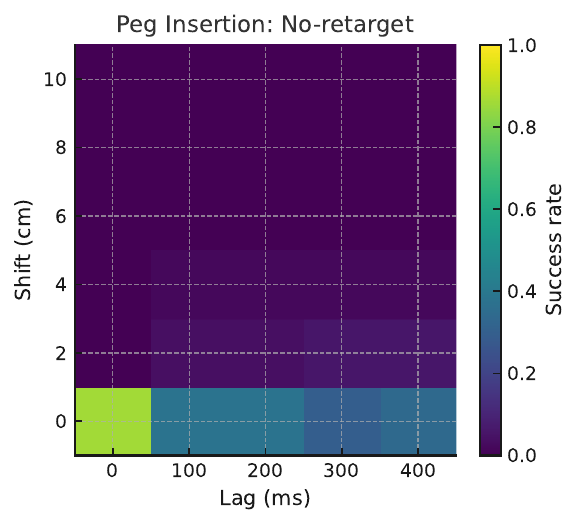} &
    \includegraphics[width=0.195\linewidth]{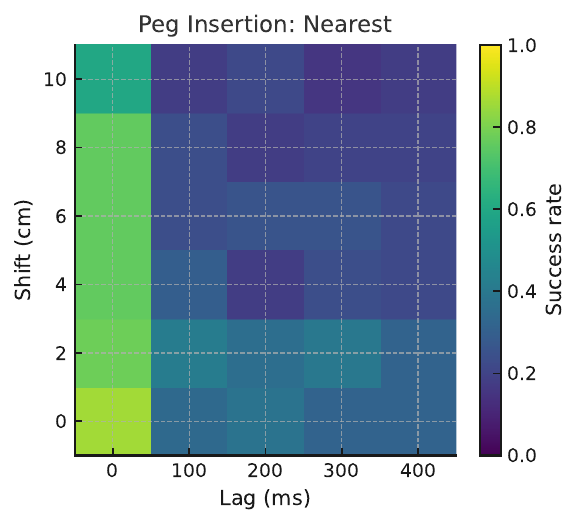} &
    \includegraphics[width=0.195\linewidth]{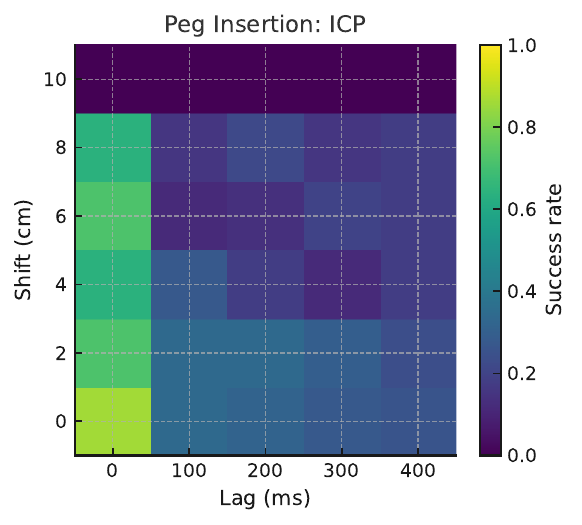} &
    \includegraphics[width=0.195\linewidth]{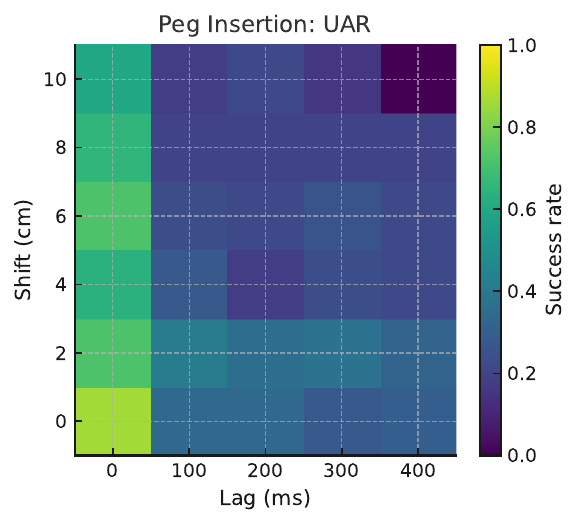} &
    \includegraphics[width=0.195\linewidth]{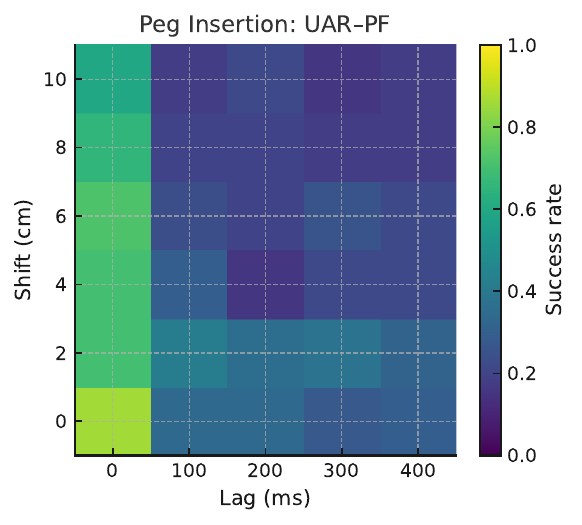}
  \end{tabular}
  \vspace{-0.35em}
  \caption{All tasks: success over Shift$\times$Lag (50 seeds/cell). Columns are baselines (No-retarget, Nearest, ICP, UAR, UAR--PF). Common color scale across all panels.}
  \label{fig:all-heatmaps}
\end{figure*}

\vspace{0.25em}
\subsection{Quantitative Results (Pick and Place)}
\label{subsec:quant-pick}

\paragraph{Headline results}
UAR--PF maintains \emph{high} success across the entire grid, including the hardest cells:
at $(10\,\mathrm{cm},\,400\,\mathrm{ms})$ it achieves $0.86$ success, versus UAR $0.10$, ICP $0.24$, Nearest $0.10$, and \emph{no-retarget} $0.00$.
At $(8\,\mathrm{cm},\,300\,\mathrm{ms})$, UAR--PF and UAR both reach $0.84$ while ICP and Nearest fall to $0.32$ and $0.08$.
At $(4\,\mathrm{cm},\,100\,\mathrm{ms})$, UAR--PF attains $0.92$ (UAR $0.92$, ICP $0.90$, Nearest $0.08$, \emph{no-retarget} $0.10$) (Fig.~\ref{fig:all-heatmaps}).

\paragraph{Mode comparison}
\emph{Nearest} collapses quickly once the object is teleported:
for $r{\ge}4$\,cm it is $\le\!0.15$ and becomes $0.00$ at $r{\ge}6$\,cm across lags.
\emph{ICP} is competitive at small shifts but fails for large teleports even at $L{=}0$\,ms (e.g., $r{=}8$\,cm: $0.10$; $r{=}10$\,cm: $0.05$).
\emph{UAR} tracks UAR--PF in easy/moderate cells but can collapse under the largest shift\,+\,lag ($r{=}10$\,cm, $L{=}400$\,ms: $0.10$), where PF’s belief smoothing prevents mis-aims.
\emph{UAR--PF} is consistently best/tied-best, staying $\ge\!0.80$ across nearly the whole grid.

\paragraph{Latency and travel}
Mean retarget latency scales with injected $L$. For UAR--PF, mean latencies are about
\SIlist{29;122;215;310;397}{\milli\second} at $L\!=\!\{0,100,200,300,400\}$.
Average EE travel is similar across modes (UAR--PF: \SI{23.0}{\meter}; ICP: \SI{22.6}{\meter};
Nearest: \SI{24.8}{\meter}; UAR: \SI{23.0}{\meter}; \emph{no-retarget}: \SI{22.4}{\meter}),
indicating robustness does not require longer paths.

\vspace{0.25em}
\subsection{Quantitative Results (Push)}
\label{subsec:quant-push}

\paragraph{Overall trends}
Retargeting helps across the grid (Fig.~\ref{fig:all-heatmaps}). Averaged over all 30 cells, success is: UAR $0.84$, UAR--PF $0.82$, ICP $0.45$, Nearest $0.33$, \emph{no-retarget} $0.23$.
Representative cells: at $(10\,\mathrm{cm},\,400\,\mathrm{ms})$ UAR $0.72$, UAR--PF $0.70$ (ICP $0.04$, Nearest $0.12$, \emph{no-retarget} $0.04$);
at $(8\,\mathrm{cm},\,300\,\mathrm{ms})$ both UAR and UAR--PF reach $0.80$ (ICP $0.06$, Nearest $0.14$, \emph{no-retarget} $0.04$);
at $(6\,\mathrm{cm},\,200\,\mathrm{ms})$ UAR--PF $0.80$ vs.\ UAR $0.78$ (ICP $0.44$, Nearest $0.22$, \emph{no-retarget} $0.02$).
In easy cases $(r{=}0,\,L{=}0)$ all methods are high (UAR $0.98$, UAR--PF $0.92$, ICP $0.76$, Nearest $0.74$, \emph{no-retarget} $0.76$).

\paragraph{Efficiency and replans}
EE travel is comparable between UAR/UAR--PF (e.g., at $(10\,\mathrm{cm},\,400\,\mathrm{ms})$: UAR \SI{20.92}{m}, UAR--PF \SI{21.24}{m}).
Retargeting modes use a budget of one re-sense/replan (means $\approx\!0.3$--$0.7$ in the hardest cells).


\vspace{0.25em}
\subsection{Retarget Latency and Replans}
\label{subsec:latency-replans}
\begin{figure}[t]
  \centering
  \subfloat[Push]{%
    \includegraphics[width=0.48\linewidth]{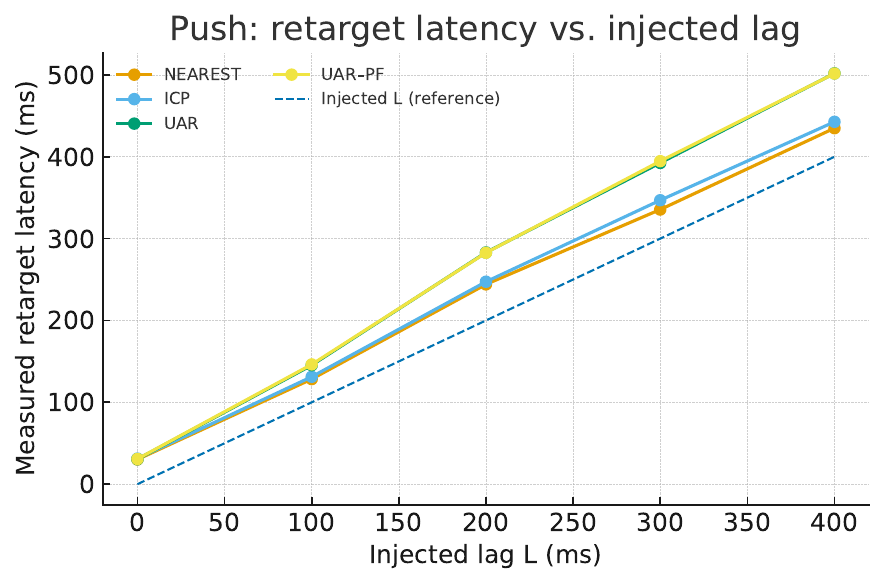}}%
  \hfill
  \subfloat[Pick]{%
    \includegraphics[width=0.48\linewidth]{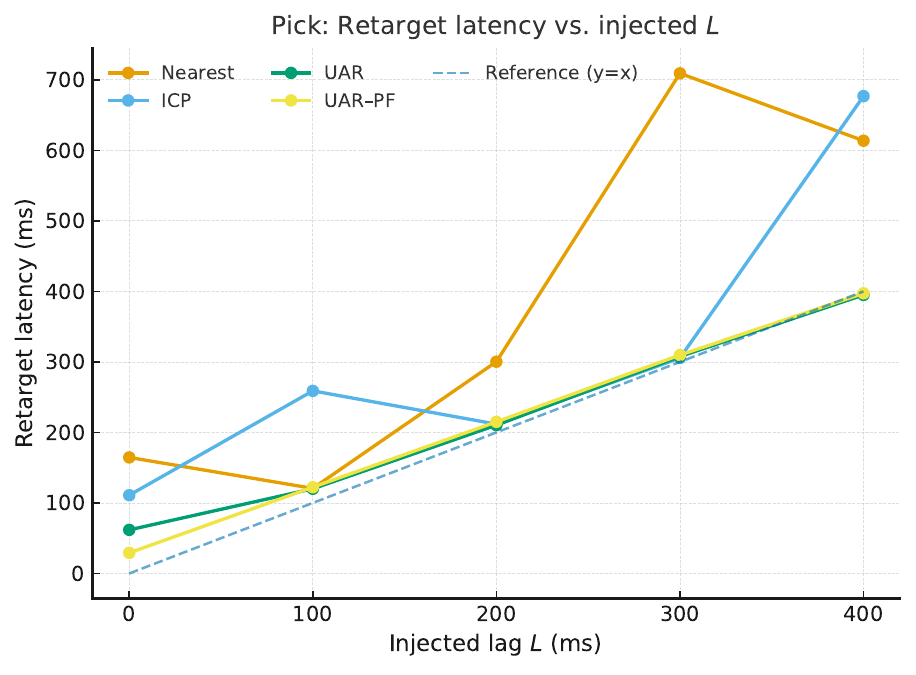}}%
  \vspace{-0.4em}
  \caption{\textbf{Retarget latency vs.\ injected $L$.} Means with std bars.
  Push tracks $L$ closely for all modes; on pick, ICP shows added overhead at high $L$.}
  \label{fig:latency}
\end{figure}

\paragraph*{Latency}
Measured latency closely tracks $L$ with small algorithmic overhead (tens of ms; see Fig.~\ref{fig:latency}). For push, means cluster around \SIlist{104;199;295;392}{\milli\second} at $L\!=\!\{100,200,300,400\}$; for $L{=}0$, ICP averages $\approx\SIrange{27}{28}{\milli\second}$, with UAR/Nearest slightly lower.

\vspace{0.25em}
\subsection{Quantitative Results (Stacking)}
\label{subsec:quant-stack}

\paragraph{Overall trends}
Stacking is strict because success requires both a correct pick and a precise place. Averaged over all $6{\times}5$ cells (50 seeds/cell), UAR--PF attains the highest stacked success ($0.773$), followed by UAR ($0.745$), ICP ($0.513$), Nearest ($0.123$), and \emph{no-retarget} ($0.250$). Hardest cell $(10\,\mathrm{cm},\,400\,\mathrm{ms})$: only UAR--PF remains nonzero ($0.64$); others drop to $0.00$. At $(8\,\mathrm{cm},\,300\,\mathrm{ms})$ UAR and UAR--PF both reach $0.72$ (ICP/Nearest/\emph{no-retarget}: $0.00$). At $(6\,\mathrm{cm},\,200\,\mathrm{ms})$ UAR/UAR--PF $0.78$ (ICP $0.66$).

\paragraph{Pick vs.\ place bottlenecks}
Across the grid, \emph{picked} rates exceed \emph{stacked}, reflecting placement sensitivity under lag (averages: UAR--PF $0.817$, UAR $0.789$, ICP $0.560$, Nearest $0.188$, \emph{no-retarget} $0.275$). In the hardest cell, UAR--PF still picks $0.68$ and stacks $0.64$, whereas UAR picks $0.06$ and stacks $0.00$.

\paragraph{Latency and travel}
Pick-phase mean latencies (ms): UAR--PF $88.5$, UAR $117.1$, ICP $238.8$, Nearest $362.3$, \emph{no-retarget} $370.7$; place-phase: UAR--PF $180.3$, UAR $211.0$, ICP $543.0$, Nearest $385.0$, \emph{no-retarget} $641.6$. Average EE travel is similar for UAR/UAR--PF ($\approx 44.4$–$45.5$ in our units); lower travel for ICP/Nearest largely reflects unfinished or early-aborted motions.

\subsection{Quantitative Results (Peg Insertion)}
\label{subsec:quant-peg}

\paragraph{Overall trends}
Insertion is the bottleneck: across the $6{\times}5$ Shift$\times$Lag grid (50 seeds/cell) the mean \emph{inserted} success falls with stress. Averaged over all cells, inserted success is: Nearest $0.297$, ICP $0.271$, UAR $0.339$, UAR--PF $0.343$, and \emph{no-retarget} $0.085$.

\paragraph{Mode comparison}
UAR--PF is best overall (34.3\%), narrowly ahead of UAR (33.9\%) and Nearest (29.7\%), with all three clearly above ICP (27.1\%) and \emph{no-retarget} (8.5\%). Nearest’s simple geometric recomputation remains surprisingly competitive at moderate stress; PF’s belief smoothing preserves UAR’s robustness under larger shift/lag when single-frame estimates become jittery. ICP degrades when pre/post-lag clouds have low overlap.

\paragraph{Latency and travel}
Retarget latency follows the injected delay with small overhead; for UAR--PF, mean latency (ms) at $L=\{0,100,200,300,400\}$ is approximately $\{29,126,235,340,448\}$; Nearest/UAR are similar, and ICP’s extra alignment cost is negligible relative to injected $L$. Mean EE travel is comparable across modes (e.g., $\approx\!30$ units), indicating higher robustness does not demand longer paths.

\begin{figure}[t!]
  \centering
  \includegraphics[
    width=0.88\columnwidth,
    height=0.22\textheight,
    keepaspectratio,
    trim=6pt 8pt 6pt 2pt,clip
  ]{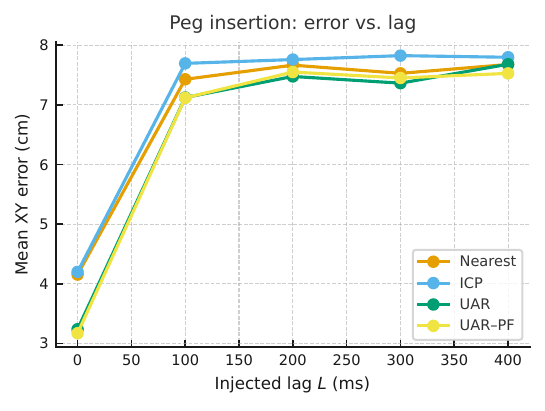}
  \vspace{-0.4em}
  \caption{\textbf{Peg insertion: XY error vs.\ injected lag.}
  Mean lateral error (cm) averaged over shifts.}
  \label{fig:peg-xyerr}
  \vspace{-0.6em}
\end{figure}

\paragraph{Accuracy (XY error)}
Mean lateral error (cm) increases with lag; for UAR--PF it is roughly $\{1.68,2.36,3.15,3.92,4.75\}$ at $L=\{0,100,200,300,400\}$ (see Fig.~\ref{fig:peg-xyerr}), with Nearest close and ICP substantially worse, mirroring the success trends.
\vspace{0.25em}
\subsection{Failure Modes}
\label{subsec:failmodes}
Typical failures include:
(i) overshoot when a teleport places the object near edges and post-overshoot lands off target;
(ii) guarded cloud resets that become overly conservative under large $L$, causing under-reach;
(iii) persistent jams against edges/fixtures;
(iv) loss of stable face contact after a partial touch.
The stall/slip monitors mitigate (iii)/(iv), while UAR--PF’s belief smoothing mitigates (ii) by stabilizing goals under noisy/guarded updates.
\section{CONCLUSIONS}

\textbf{U{-}LAG} is a lightweight goal-retargeting layer that updates pre-, contact-, and post-waypoints online under delayed/noisy perception while leaving the low-level controller unchanged. We instantiate four retargeters (Nearest, ICP, UAR, and UAR--PF) with simple reliability hooks for stable contact. On a unified Shift$\times$Lag stress test in PyBullet/PandaGym across \emph{push}, \emph{pick}, \emph{stacking}, and \emph{peg insertion}, all retargeters substantially outperform a no-retarget baseline. UAR--PF is most robust in the hardest regimes, maintaining high pick/stack rates and delivering the best insertion success, while keeping end-effector travel comparable to other modes and adding only tens of milliseconds of compute overhead. Future work includes hardware deployment, extensions to mobile manipulation, and learned uncertainty models for online retargeter selection.

\addtolength{\textheight}{-12cm}   









\end{document}